\documentclass[letterpaper, 10 pt, conference]{ieeeconf}  

\IEEEoverridecommandlockouts                              

\usepackage{graphicx} 
\usepackage{times} 
\usepackage{amsmath} 
\usepackage{amssymb}  
\usepackage{adjustbox}
\let\labelindent\relax

\usepackage{mathtools}
\usepackage{ulem}
\usepackage{threeparttable}
\usepackage{booktabs}
\usepackage{graphicx}
\usepackage{wrapfig}
\usepackage{multicol}
\usepackage{multirow}
\usepackage{cite}
\usepackage{bm}
\usepackage{algorithmic}
\usepackage[ruled]{algorithm2e}

\SetCommentSty{mycommfont}
\usepackage{subcaption}
\usepackage{eqlist}
\usepackage{amsfonts}
\usepackage{enumitem}
\usepackage{lipsum}
\usepackage[usenames,dvipsnames]{xcolor}
\usepackage{comment}

\newcommand{\etal}{\textit{et~al.}}

\usepackage{multirow}

\usepackage{tikz}
\usetikzlibrary{fit,shapes.misc}

\title{Robotic Paper Wrapping by Learning Force Control}
\author{Hiroki Hanai$^{1}$, Takuya Kiyokawa$^{1}$, Weiwei Wan$^{1}$, and Kensuke Harada$^{1}$
\thanks{$^{1}$Graduate School of Engineering Science, Osaka University, Japan.}
\thanks{Contact: Kensuke Harada, {\tt\small harada@sys.es.osaka-u.ac.jp}}}

\begin{document}

\maketitle
\thispagestyle{empty}
\pagestyle{empty}

\begin{abstract} 

Robotic packaging using wrapping paper poses significant challenges due to the material’s complex deformation properties. The packaging process itself involves multiple steps, primarily categorized as folding the paper or creating creases. Small deviations in the robot's arm trajectory or force vector can lead to tearing or wrinkling of the paper, exacerbated by the variability in material properties.

This study introduces a novel framework that combines imitation learning and reinforcement learning to enable a robot to perform each step of the packaging process efficiently. The framework allows the robot to follow approximate trajectories of the tool-center point (TCP) based on human demonstrations while optimizing force control parameters to prevent tearing or wrinkling, even with variable wrapping paper materials.

The proposed method was validated through ablation studies, which demonstrated successful task completion with a significant reduction in tear and wrinkle rates. Furthermore, the force control strategy proved to be adaptable across different wrapping paper materials and robust against variations in the size of the target object.

\end{abstract}

\section{Introduction}
\label{sec:introduction}
Automation in product packaging is increasingly in demand as the final stage in the manufacturing process. However, robotic wrapping using wrapping paper remains an unresolved challenge, despite its potential benefits.

\begin{figure}[t]
    \begin{minipage}[b]{\hsize}
    \centering
    \includegraphics[width=\linewidth]{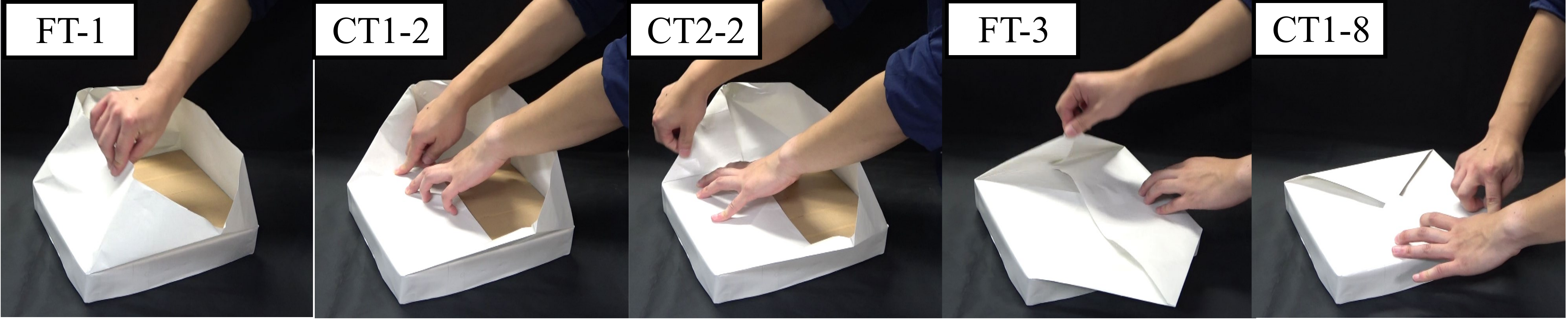}
    \subcaption{\small{Human demonstration}}
    \label{fig:wrapping_demo}
    \end{minipage}\\
    \begin{minipage}[b]{\hsize}
    \centering
    \includegraphics[width=\linewidth]{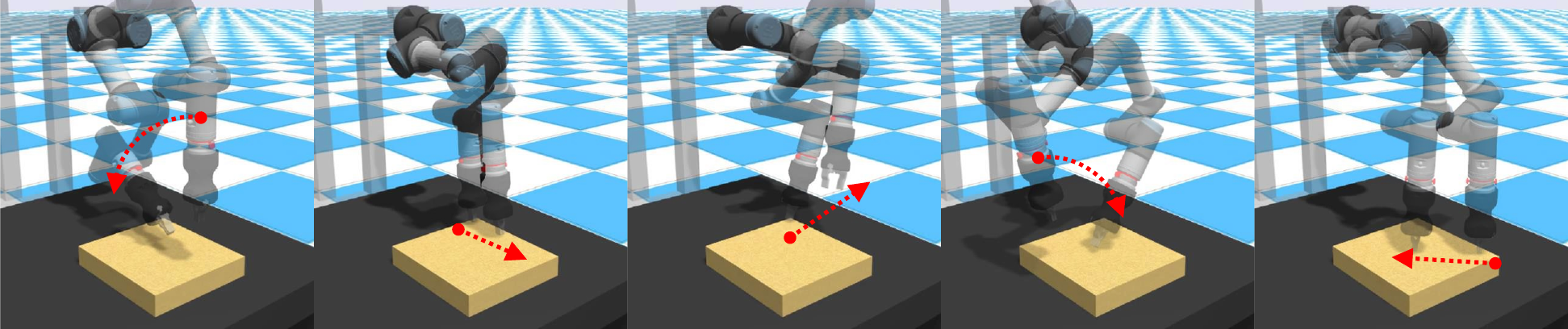}
    \subcaption{\small{Trajectory learning in simulation}}
    \label{fig:all_wrapping_operation_sim}
    \end{minipage}\\
    \begin{minipage}[b]{\hsize}
    \centering
    \includegraphics[width=\linewidth]{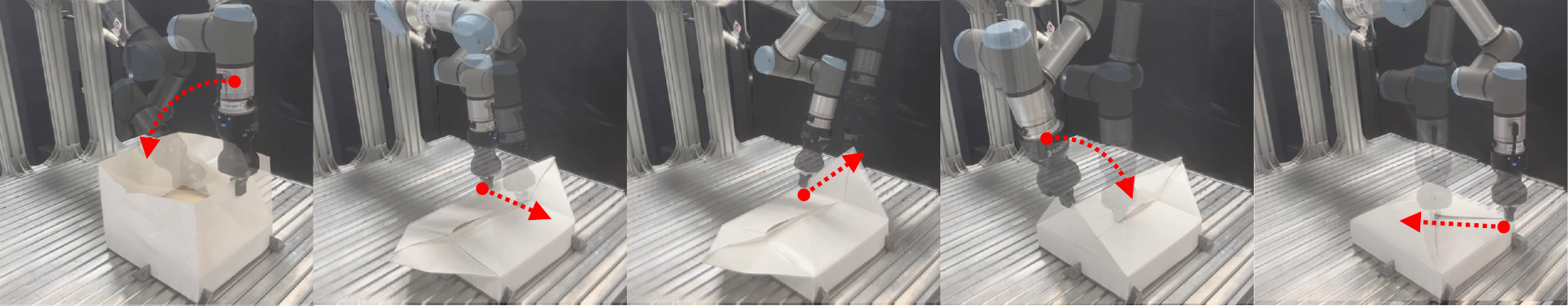}
    \subcaption{\small{Force learning in real-world}}
    \label{fig:all_wrapping_operation_middle}
    \end{minipage}
    \caption{\small{Wrapping actions addressed in our learning framework}}
\end{figure}

Consider the task of wrapping a box-shaped object. As observed in human wrapping tasks (illustrated in Fig. \ref{fig:wrapping_demo}), this process consists of three main sub-tasks: folding the paper along the box's edges, labeled as the folding task (FT); pressing the paper onto the surface to create creases, labeled as the creasing task 1 (CT1); and creasing the paper's mountain folds without contacting the box, labeled as the creasing task 2 (CT2). For successful automation, a robot must be capable of handling diverse paper materials with varying degrees of tearability and susceptibility to wrinkling. Even small deviations in the robot's end-effector trajectory or contact force during folding or creasing can result in tears or wrinkles, making precise force control a critical factor. These control parameters must be dynamically adjusted to account for the properties of the paper material, which makes this task both complex and non-trivial.

 
To address this problem, we propose a learning framework for autonomous wrapping that integrates imitation learning (IL)~\cite{9560942} and reinforcement learning (RL)~\cite{ibarz2021train}. 
In this framework, IL allows the robot to learn approximate trajectories of the tool-center point (TCP) from human demonstrations, resulting in smooth and stable motion. Although force control is effective in wrapping tasks to prevent the paper from tearing and wrinkling, the force control policies obtained through IL may not be directly transferred to a robotic system due to differences in the dynamic characteristics between humans and robots.
RL, on the other hand, enables the robot to acquire appropriate trajectories and force control parameters stably interacting with the environment. Despite its advantages, RL may result in paper tearing and wrinkling due to inefficient policy sampling during learning and repeated trial and error. To overcome these limitations, we propose a hybrid approach that combines the strengths of IL and RL. By using IL to constrain the search space of RL with pre-learned trajectories, we mitigate the inefficiencies inherent in RL's exploration. Simultaneously, RL refines force control parameters, complementing the limitations of IL.  In our approach, IL first identifies task phases and extracts target trajectories from human wrapping demonstrations. Subsequently, RL learns the appropriate force control strategies, generating appropriate force vectors tailored to the material properties of the wrapping paper.
This integrated learning framework enables the robot to acquire both trajectory and force control parameters, facilitating the precise application of force in the right amount and direction for different paper materials. As a result, our method offers a robust and efficient solution for performing reliable wrapping operations across a wide range of materials.

This paper is organized as follows: after introducing related works in Section 2, the problem setting is explained in Section 3. The proposed method is explained in Section 4. Finally, we show the experimental results to confirm the effectiveness of the proposed method in Section 5. 

\section{Related Works}

For the research on deformable object manipulation, 
much research on cloth manipulation has been done so far. 
The research includes online cloth folding with shape estimation~\cite{tanaka2021}, 
estimation of the tensile vector for stretching cloth~\cite{seita2020deep}, 
simulation-based state estimation~\cite{wang2022learning}, 
predicting the cloth state using graph neural networks~\cite{9981669}, 
and predicting the trajectory of cloth using reinforcement learning~\cite{9196659}.
However, these methods do not take into account the stiffness of the material and are not suitable for folding or creating creases in paper since we cannot prevent paper from tearing or wrinkling. 

The research on rope manipulation includes the prediction model from image-based demonstration data~\cite{7989247}, 
estimation of physical parameters~\cite{10093017}, 
and realization of tying task by using the image-based imitation learning~\cite{9197121}.
In addition, some research on paper manipulation has been done, including paper folding by predicting the force generated on paper~\cite{choi2023deep}, continuous valley folding of origami with a multi-fingered~\cite{namiki2015robotic}, and vision-based paper folding by predicting paper deformation ~\cite{6651522}.
However, there is no research on the full automation of the packaging operation without the use of a specialized hand for paper folding.

For the research on RL of deformation-adaptive controllers, Zhao~\etal~\cite{zhao2022offline} introduced a framework that combines offline meta-RL and online fine-tuning to achieve various shapes of peg insertion tasks with high success rates with fixed material and friction.
Similarly, Pong~\etal~\cite{pong2022offline} proposed a learning framework that combines offline meta-RL and estimation of the reward function with online fine-tuning. 

For the research on few-shot IL from demonstrations, 
Du~\etal~\cite{du2023behavior} presented a state and action retrieval IL approach to estimate current state and action strategies from a large unlabeled dataset. 
Jang~\etal~\cite{jang2021bc} proposed a method to acquire action strategies, such as trajectory and gripper opening/closing, of a robot to perform tasks in zero-shot by combining a large multi-task dataset in image format, a video of a human demonstration, and natural language vectors. 
However, it is unclear whether the framework can be adapted to flexible object manipulation, which needs careful adjustment of force control parameters. 

For the research on combining RL and IL, 
Wang~\etal~\cite{wang2022adaptive} used IL to acquire trajectories from human demonstrations and RL to learn force control strategies that adapt to the trajectories obtained through IL. 
However, the aforementioned approach cannot be implemented to the wrapping task. On the other hand, this is the first trial on realizing the wrapping task needs careful adjustment of force control parameters comprising both folding and creating creases. 


Recently, some vision-based imitation learning like Diffusion Policy \cite{chi2023diffusionpolicy} and Action Chunking Transformer\cite{zhao2023learningfinegrainedbimanualmanipulation}
has been proposed and is widely used. Although we do not use such recent imitation learning, we believe that we can use these imitation learning algorithm in combination with the reinforcement learning to train the force control parameters and efficiently realize the wrapping task. 


\section{Wrapping Task Setting}

We consider the wrapping task of a rectangular-shaped object. 
The wrapping process comprises multiple phases, which can be divided into four folding tasks (FT), eight creasing tasks that press the paper against the box surface (CT1), and four creasing tasks for forming mountain folds without contacting the box (CT2). Each task requires specific trajectory and force control strategies to avoid tearing or wrinkling the paper.

\begin{figure}
    \centering
    \includegraphics[width=\linewidth]{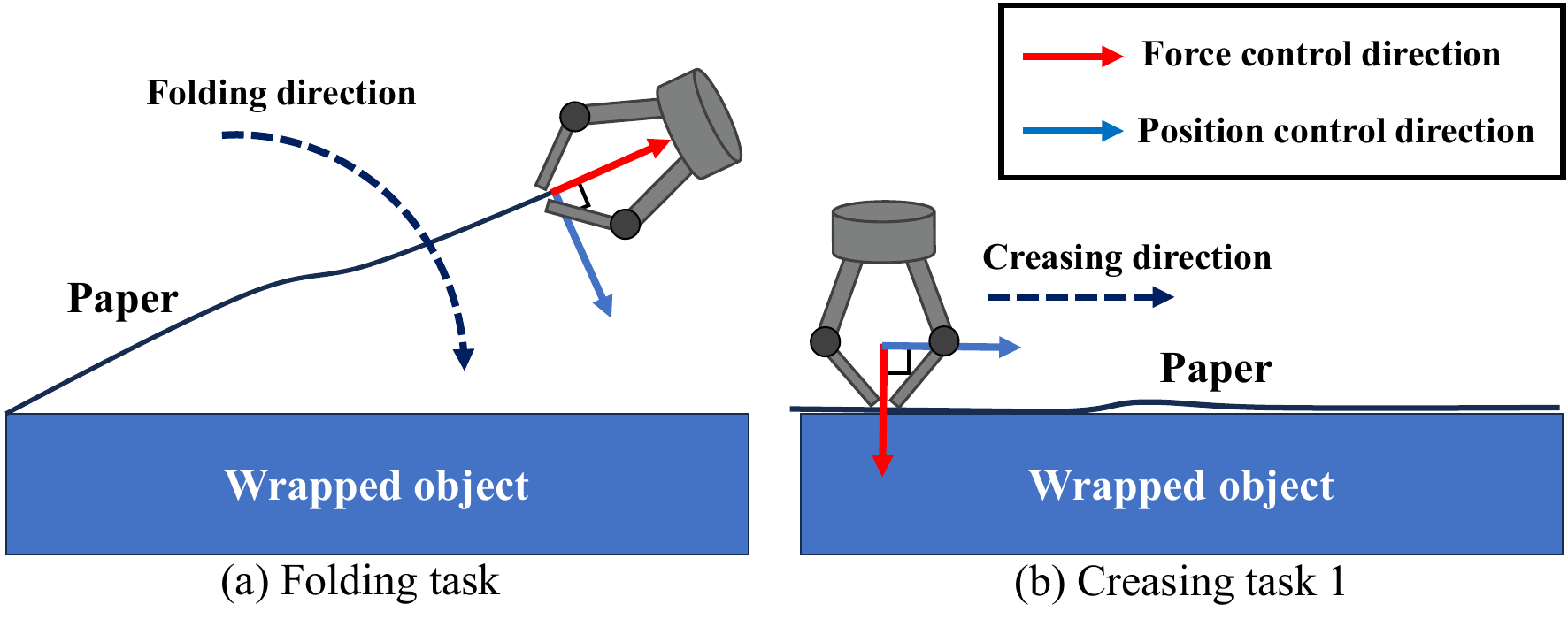}
    \caption{\small{Hybrid position/force control for wrapping tasks}}
    \vspace{-3mm}
    \label{fig:FT_CT1_pd_direction}
\end{figure}

\noindent
\textbf{FT:} 
Tears are prone to occur when the paper is misaligned during folding, while wrinkles typically form when excessive force is applied. To mitigate these issues, the paper must be stretched with an appropriate amount of force during the folding process. As shown in Fig.~\ref{fig:FT_CT1_pd_direction}(a), force control is needed in the direction parallel to the paper surface, whereas trajectory control is needed in the direction perpendicular to it.

\noindent
\textbf{CT1:} During this phase, excessive force can cause both tearing and wrinkling of the paper. On the other hand, insufficient force may prevent the crease from forming correctly. Therefore, trajectory control should be implemented in the tangential direction to the paper surface, while force control should be applied in the normal direction, as illustrated in Fig.~\ref{fig:FT_CT1_pd_direction}(b).

\noindent
\textbf{CT2:} In this phase, precise trajectory control is crucial to ensure the gripper follows the creased section of the paper accurately. Wrinkles tend to form when the gripper opens or closes, causing the paper to deflect upon contact with the finger. However, if the gripper is positioned too far from the paper, it will fail to create a proper crease. Therefore, precise position control is vital to maintain the gripper’s alignment with the creased portion of the paper.

\section{Wrapping Skill-Learning Methodology}
\subsection{Overview}
\begin{figure}
    \centering
    \includegraphics[width=\linewidth]{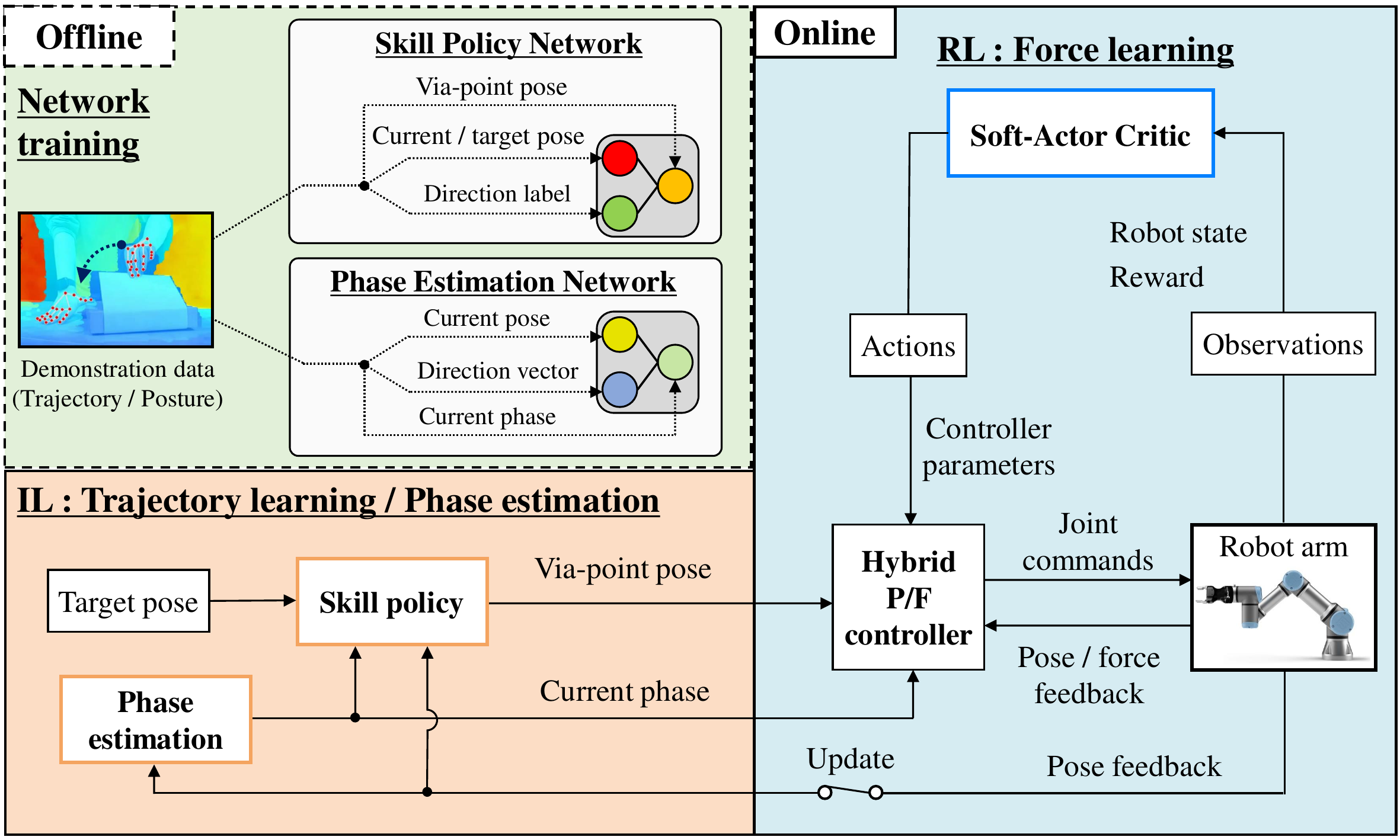}
    \caption{\small{Overview of the proposed learning method and its execution pipeline}}
        \vspace{-3mm}
    \label{fig:overview_of_the_proposed_method}
\end{figure}

A schematic of the proposed system combining the IL and RL is shown in Fig. ~\ref{fig:overview_of_the_proposed_method} and in the pseudo-code of Algorithm \ref{alg_ILRL_learning} . The system was divided into two components: IL and RL.
The IL part estimates the current phase by using the phase estimation network and generates the target position of TCP by using the skill policy network. 
At each time step, the phase estimation network estimates the current phase 
from TCP's current position. 
In the skill policy network, an appropriate policy is selected according to the estimated phase, and the current and the target position 
of the TCP. 
On the other hand, by using RL, the agent learns the parameters of hybrid position/force control \cite{hybrid} to realize the target trajectory in each phase according to the paper material.
\begin{algorithm}[t]
\caption{Learning system of combined IL/RL}
\label{alg_ILRL_learning}
\SetKwInOut{Input}{Initialization}
\SetKw{Processing}{Processing}
\SetKwData{F}{F}
\SetKwData{Cf}{C1}
\SetKwData{Cl}{C2}
\SetKwData{Task}{task}
\SetKwData{Taskbar}{task-}
\SetKwData{Cm}{Control\_method}
\SetKwData{IK}{IK\_solver}
\SetKwData{mod}{mod}
\SetKwComment{tcc}{}{}
\SetKwIF{Ifcon}{ElseIf}{Else}{if}{then continue}{else if}{else}{}
\SetKwIF{Ifbre}{ElseIf}{Else}{if}{then break}{else if}{else}{}
\Input{\\}
initial position $\bm{p}_{\text{init}}$ and goal position $\bm{p}_{\text{g}}$\;
update steps of via-point position $n_{\text{IL}}$
reference vector for labeling $\bm{l}_{\text{d}}^{\text{ref}}$

\Processing{\\}
\For{$n=1$ \KwTo $N$ episodes}{
    \For{$t=1$ \KwTo $T$ steps}{
        Get current position $\bm{p}_{\text{c}}$ and force $\bm{f}_c$\;
        {\color{blue}\tcc{/* IL part (offline policy) */}}
        \If{$t$ mod $n_{\text{IL}}$}{
            $\bm{v} \leftarrow \bm{p}_{\text{g}} - \bm{p}_{\text{init}}$\; 
            $\bm{l}_{\text{d}} \leftarrow \underset{\bm{l}_{\text{d}}^{\text{ref}}[i]}{\operatorname{argmax}} \cos{\left( \bm{v}, \bm{l}_{\text{d}}^{\text{ref}} \right)}$\;
            Estimate current phase $ph_{\text{c}}$ from phase estimation policy $\pi(\bm{ph}| \bm{p}_{\text{c}}, \bm{v})$\; 
            Get via-point position $\bm{p}_{\text{v}}$ from skill policy $\pi_{ph_{\text{c}}}(\bm{p}_{\text{v}}|\bm{p}_{\text{c}},\bm{p}_{\text{g}}, \bm{l}_{\text{d}})$\;
        }

        {\color{blue}\tcc{/* RL part (online policy) */}}
        $\bm{p}_{\text{e}} \leftarrow \bm{p}_{\text{v}} - \bm{p}_{\text{c}}$\;
        $\bm{f}_{\text{e}} \leftarrow \bm{f}_{\text{c}} - \bm{f}_{\text{idl}}$\;
        Get observation $\bm{o}_t=[\bm{p}_{\text{e}},\dot{\bm{p}}_{\text{c}},\bm{f}_{\text{c}}, \bm{a}_{t-1}]$\;
        Compute policy action $\pi_{\theta}(\bm{a}_{t}|\bm{o}_t)$\;
        $\mathbf{cmd} \leftarrow \Cm (\bm{a}_{t}, \bm{p}_{\text{e}}, \bm{f}_{\text{e}})$\;
        $\bm{q} \leftarrow \IK (\mathbf{cmd})$\;
        Compute reward function $r(\bm{o}, \bm{a})$\;
        \Ifcon{$\bm{q}$ not exists or $\|d\bm{q}/dt\| > \dot{\bm{q}}_{\text{max}}$}{}
        \Ifbre{$\bm{f}_{\text{c}} > \bm{f}_{\text{max}}$}{}
        Acurate $\bm{q}$ on robot\;
    }
    Reset to $\bm{p}_{\text{init}}$\;
}
\end{algorithm}

\subsection{IL for Nominal Trajectory and Phase Estimation}

We use IL to learn the nominal trajectory, facilitating stable and rapid learning for the RL-based controller. This approach improves sample efficiency by narrowing the RL search space, reducing instances of paper tearing or wrinkling caused by repeated trial-and-error during the learning process. In the IL phase, the skill policy—specifically, the nominal trajectory of the TCP is learned from human demonstrations of the wrapping task.

However, the nominal trajectory required varies across different phases. To cope with this problem, a phase estimation network is employed to identify the current phase, allowing the system to switch between three different skill policy networks based on the estimated phase. This enables efficient learning of the appropriate skill policy for each phase.

\subsubsection{Generating Datasets from Human Demonstrations}

We prepare a demonstration dataset $\mathcal{D} = \left\{ \tau^1, \tau^2, \ldots, \tau^N \right\}$ from human demonstrations of each phase in wrapping task, where each data $\tau^i$ is composed of time series of the TCP position. 
By using Mediapipe\cite{mediapipe}, we obtain the motion of the fingertip. In FT and CT2, the midpoint of the fingertips of the thumb and index finger is assumed to be the TCP. In CT1, the tip of the index finger is assumed to be the TCP. 

Data expansion is performed by using the data relabeling\cite{lynch2020learning} where the starting and ending points are randomly changed to obtain various current pose $\bm{p}_{\text{c}}$, via-point pose $\bm{p}_{\text{v}}$, and target pose $\bm{p}_{\text{g}}$ of the TCP. 
Given the vectors $\bm{v}$ representing the motion direction of the TCP and 
$\bm{ph}$ representing the current phase, we can form the dataset for training the phase estimation network $\mathcal{D}_{\text{ph}} = \left\{ \bm{d} \hspace{1mm} | \hspace{1mm} \bm{d} = \left( \bm{p}_{\text{c}},\bm{v},\bm{ph} \right) \right\}$ and the dataset for training the skill policy network for each task $\mathcal{D}_{\text{sk}}^{\text{task}} = \left\{ \bm{d} \hspace{1mm} | \hspace{1mm } \bm{d} = \left( \bm{p}_{\text{c}}^{\text{task}},\bm{p}_{\text{v}}^{\text{task}},\bm{p}_{\text{g}}^{\text{task}}, \bm{l}_{\text{d}}^{\text{task}} \right)\right\}, \text{task} = \text{F}, \text{C1}, \text{C2}$. 

\subsubsection{Phase Estimation}

\noindent
In the phase estimation network, the policy $\pi(\bm{ph} \hspace{1mm} | \hspace{1mm} \bm{p}_{\text{c}}, \bm{v})$ is trained by using the dataset $\mathcal{D}_{\text{ph}}$. The output is the phase probability. By estimating the phase, we can select an appropriate skill policy and position/force control.

We use a fully-connected and multi-input neural network: the input is the current pose of the EEF with a 128 node $\times$ two-layer structure and the direction vector with a similar structure. Then, it has a connection layer and a 256 node $\times$ two-layer structure. The output has three nodes. 
We use ReLU for the activation, Adam for the optimization, and MSE (Mean Squared Error) for the objective functions.

\subsubsection{Skill Policy}

\noindent
By using the dataset $\mathcal{D}_{\text{sk}}^{\text{task}}$ in each phase, 
we train an NN of the skill policy 
$\pi_{\text{task}} \left( \bm{p}_{\text{v}}^{\text{task}} \hspace{1mm} | \hspace{1mm} \bm{p}_{c}^{\text{task}}, \bm{p}_{\text{g}}^{\text{task}}, \bm{l}_{\text{d}}^{\text{task}} \right)$
that outputs the via-point position $\bm{p}_{\text{v}}^{\text{task}}$ from 
current pose $\bm{p}_{c}^{\text{task}}$, goal pose $\bm{p}_{\text{g}}^{\text{task}}$ and direction label $\bm{l}_{\text{d}}$ in each phase. 
By using the skill policy network, a rough trajectory of the TCP can be obtained in each phase of the wrapping task, contributing to stable control without tearing or wrinkling the wrapping paper.

Similar to the phase estimation network, we use a fully-connected and multi-input neural network: 
the input is the current and the target pose of the EEF with a 128 node $\times$ two-layer structure 
and the direction vector with a 128 node $\times$ one-layer structure. Then, it has a connection layer and a 256 node $\times$ two-layer structure. The output has seven nodes. 
We use ReLU for the activation, Adam for the optimization, and MSE (Mean Squared Error) for the objective functions.

\subsection{RL for Force Control}

\noindent
For realizing complex wrapping tasks, we use a hybrid position/force controller trained with RL. 
We use the Soft Actor-Critic for the RL agent where its outputs are the position gain, force gain, and a target force. 

\subsubsection{Desired Force}

\noindent
To fold the paper by applying the appropriate force in FT and to crease it with the appropriate force in CT1, we applied the hybrid position/force control in which the direction to control the TCP’s position is orthogonal to the direction to control the force applied by the finger. We explain how to determine the desired force used in the hybrid position/force control. The desired force is applied along the plane $P$ normal to the velocity vector of the TCP\cite{yoshikawahybrid}.
We also assume the plane $Q$ includes three TCP positions: the initial position, the current position, and the position at the one-time step before. The desired force is applied along the intersection between the planes P and Q. Combined with the amount of target force $f_r$ acquired through RL, we can define the desired force $\bm{f}_{\text{idl}}$ as

\begin{equation}
\label{eqn: idl_force}
    \bm{f}_{\text{idl}} = \dfrac{\bm{n}_{\text{Q}} \times \bm{n}_{\text{P}} }{\left\| \bm{n}_{\text{Q}} \times \bm{n}_{\text{P}} \right\|} f_r,
\end{equation}

\noindent
where $\bm{n}_P$ and $\bm{n}_Q$ denote the normal vectors of the planes $P$ and $Q$, respectively. 

\subsubsection{RL-Based Controller}
\begin{figure}
    \centering
    \includegraphics[width=\linewidth]{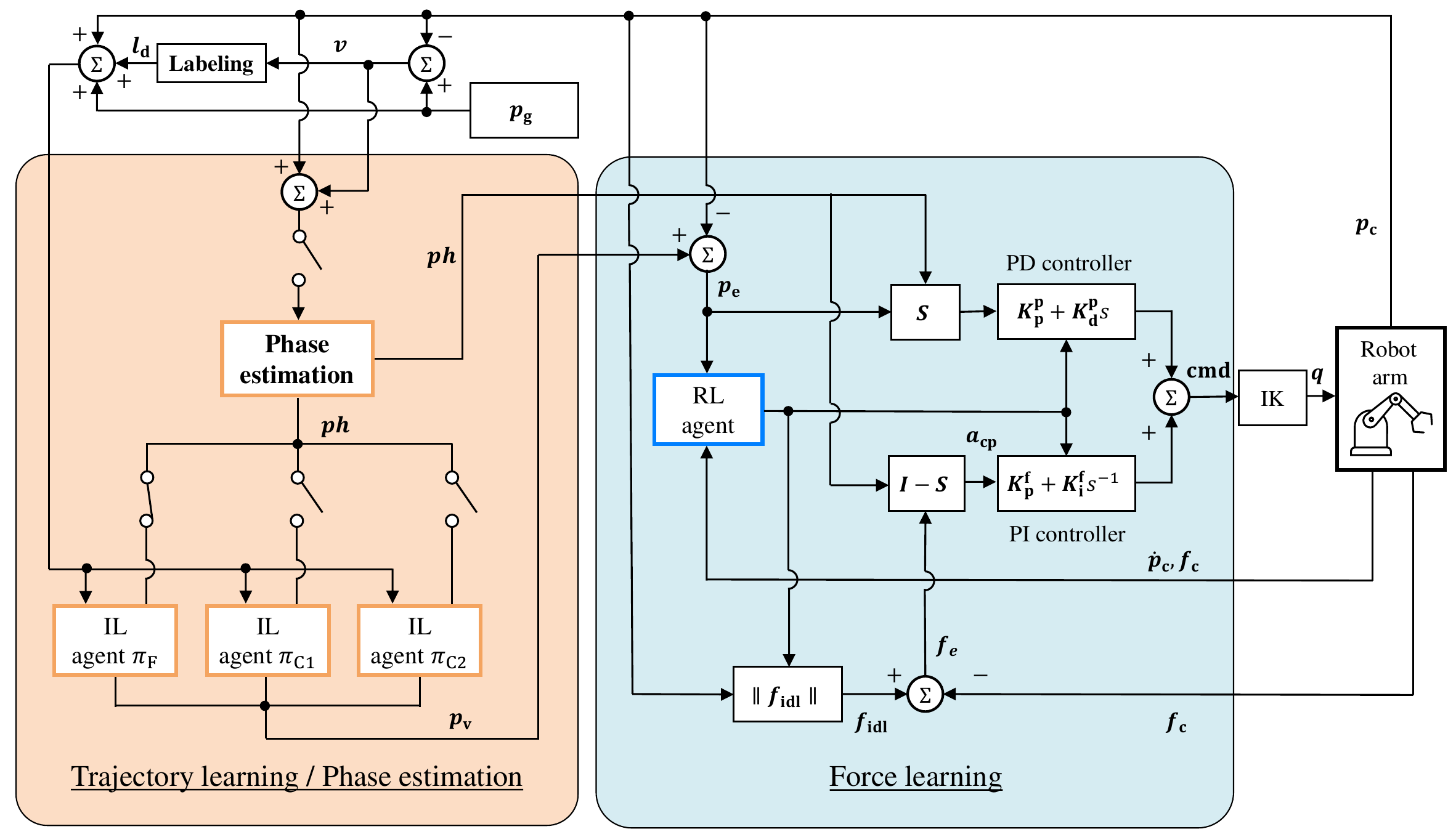}
    \caption{\small{Calculation procedure on parameters of hybrid position/force control.}}
        \vspace{-3mm}
    \label{fig:details_of_the_proposed_method}
\end{figure}

Fig.~\ref{fig:details_of_the_proposed_method} shows the diagram of the RL part. As shown in the figure, the parameters of the hybrid position/force controller are tuned according to the interactions with the environment. The nominal trajectory of the TCP's position generated in IL is the input to the RL at each time step. The position error of the TCP and the current measured force serve as feedback to both the RL agent and the hybrid position/force controller. 
The hybrid position/force controller generates the following command.

\begin{align}
    \mathbf{cmd} &= \bm{S} \left( \bm{K}_{\text{p}}^{\text{p}} \bm{p}_{\text{e}} + \bm{K}_{\text{d}}^{\text{p}} \bm{\dot{p}}_{\text{e}} \right) + \left( \bm{I} - \bm{S} \right) ( \bm{K}_{\text{p}}^{\text{f}} \bm{f}_{\text{e}} + \bm{K}_{\text{i}}^{\text{f}} \int \bm{f}_{\text{e}} dt ), \nonumber
\end{align}

\noindent
where $\bm{p}_{\text{e}}$ and $\bm{f}_{\text{e}}$ denote the error of the TCP's position and applied force. 
The hybrid position/force controller control law consists of a proportional-derivative (PD) controller for the position, a proportional-integral (PI) controller for force, and a selection matrix $\bm{S}$ that selects the appropriate contribution to the control law in the task coordinate where it is estimated by using the algorithm \cite{yoshikawahybrid}. The controllable parameters are $\bm{a}_{\text{cp}}=[\bm{K}_{\text{p}}^{\text{p}}, \bm{K}_{\text{p}}^{\text{f}}, f_r] \in \mathbb{R}^{13}$. Each parameter is self-tuned during the learning process of the RL agent.

\subsubsection{Algorithm and Reward}

In order to obtain the position/force control policy, we use Soft-Actor-Critic (SAC)~\cite{sac} as the RL algorithm; SAC is a state-of-the-art model-free and off-policy Actor-Critic deep RL algorithm based on the maximum entropy RL framework.
As an off-policy algorithm, we use a distributed prioritized experience replay approach for sample-efficient learning. The SAC implementation is based on the TF2RL repository. The reward function is defined as
\begin{align}
    r(\bm{s},\bm{a}) =& \omega_{1} H \left( \dfrac{\left\|\bm{p}_{\text{eg}}\right\|}{p_{\text{max}}}  \right) +  \omega_{2} H \left(  \dfrac{\left\|\bm{f}_{\text{e}}\right\|}{f_{\text{max}}}  \right) + \omega_{3} H \left(  \dfrac{\left\|\Delta \bm{f}_{\text{idl}}\right\|}{f_{\text{range}}}  \right) \nonumber \\
    &+ \eta, \nonumber
\end{align}
where $p_{\text{max}}$,  $f_{\text{max}}$ and $f_{\text{range}}$ are the maximum values of $\left\|\bm{p}_{\text{eg}}\right\|$, $\left\|\bm{f}_{\text{e}}\right\|$ and $\left\|\Delta \bm{f}_{\text{idl}}\right\|$, respectively, and $y=H(x)$ is a linear mapping function from $x \in [0,1]$ to $y \in [1,0 ]$.
The first term aims to acquire the TCP approaching to the target position while the second term aims to acquire the appropriate force of pulling or holding down the wrapping paper. In addition, the third term aims to avoid the paper from tearing and wrinkling due to large changes in $\bm{f}_{\text{idl}}$. $\eta$ becomes positive for successful completion of the task considering the number of steps it took to succeed, and becomes negative for tearing paper or IK cannot be solved or exceeding the joint velocity limit. 
To address safety concerns during the learning process, we check the IK solvability, joint velocity and force norm in each action. 


\section{Experiments} \label{experiments_results}


We developed a simulated environment using the Gazebo 9 simulator and a real hardware experimental setup. We use a UR-3e robotic arm which has a force/torque sensor and Robotiq Hand-e parallel gripper, both mounted on the wrist. The learning process is executed on a PC with a GeForce RTX 2080 GPU and an Intel Core i9-9980 CPU. The policy control frequency is $20\,\mathrm{Hz}$. A rectangular box with the size $0,23\times 0.27 \times 0.5\,\mathrm{m}$ was used as the object to be wrapped, with three distinct types of thickness utilized as wrapping paper.

\subsection{Training}

The RL agent was trained in both simulated and real environments. In the simulation environment, the agents adjust the position control parameter $\bm{K}_{\text{p}}^{\text{p}}$ to achieve the nominal trajectory acquired through imitation learning. In the simulation environment, agents learned $50$k time steps for $4$ subtasks 
of FT, $30$k time steps for $8$ subtasks 
of CT1, and $30$k time steps for $4$ subtasks 
of CT2.

After learning in the simulation environment, the agent is trained in the real environment to learn force control parameters 
using the real paper with $0.09\,\mathrm{mm}$ thickness through transfer learning using policies obtained in the simulation. In the real environment, the agent learned $40$k time steps for $4$ subtasks 
of FT, $30$k time steps for $8$ subtasks 
of CT1, and $30$k time steps for $4$ subtask 
of CT2. We set the condition for task accomplishment for the position error less than $10\,\mathrm{mm}$ and the rotation error less than $0.17\,\mathrm{rad}$. 

\begin{figure*}[t]
    \begin{minipage}[b]{\hsize}
    \centering
    \includegraphics[width=\linewidth]{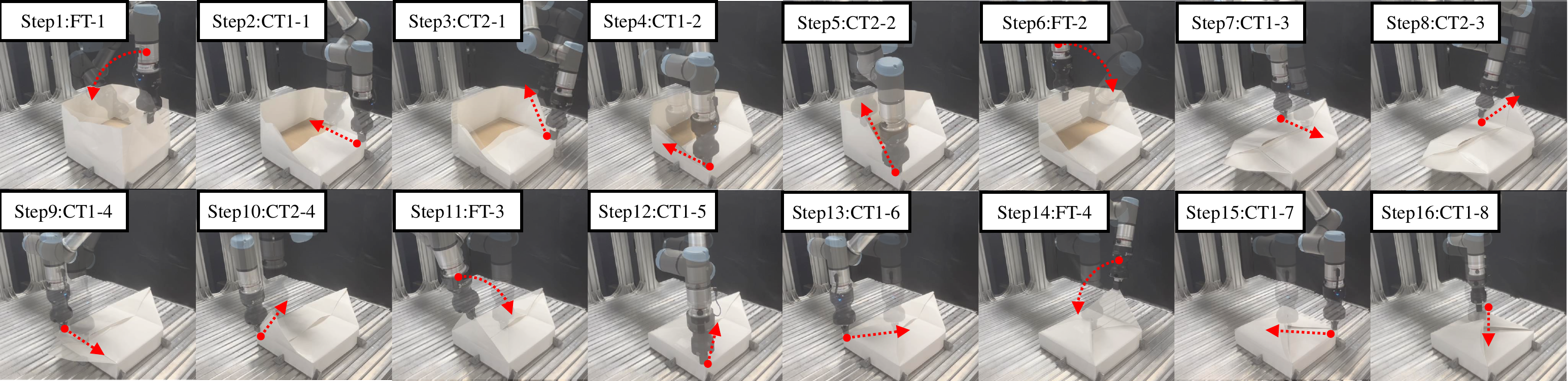}
    \subcaption{\small{Wrapping task using medium-thickness paper}}
    \label{fig:wrapping_flow_thickness_middle}
    \end{minipage}\\
    \begin{minipage}[b]{0.5\hsize}
    \centering
    \includegraphics[width=\linewidth]{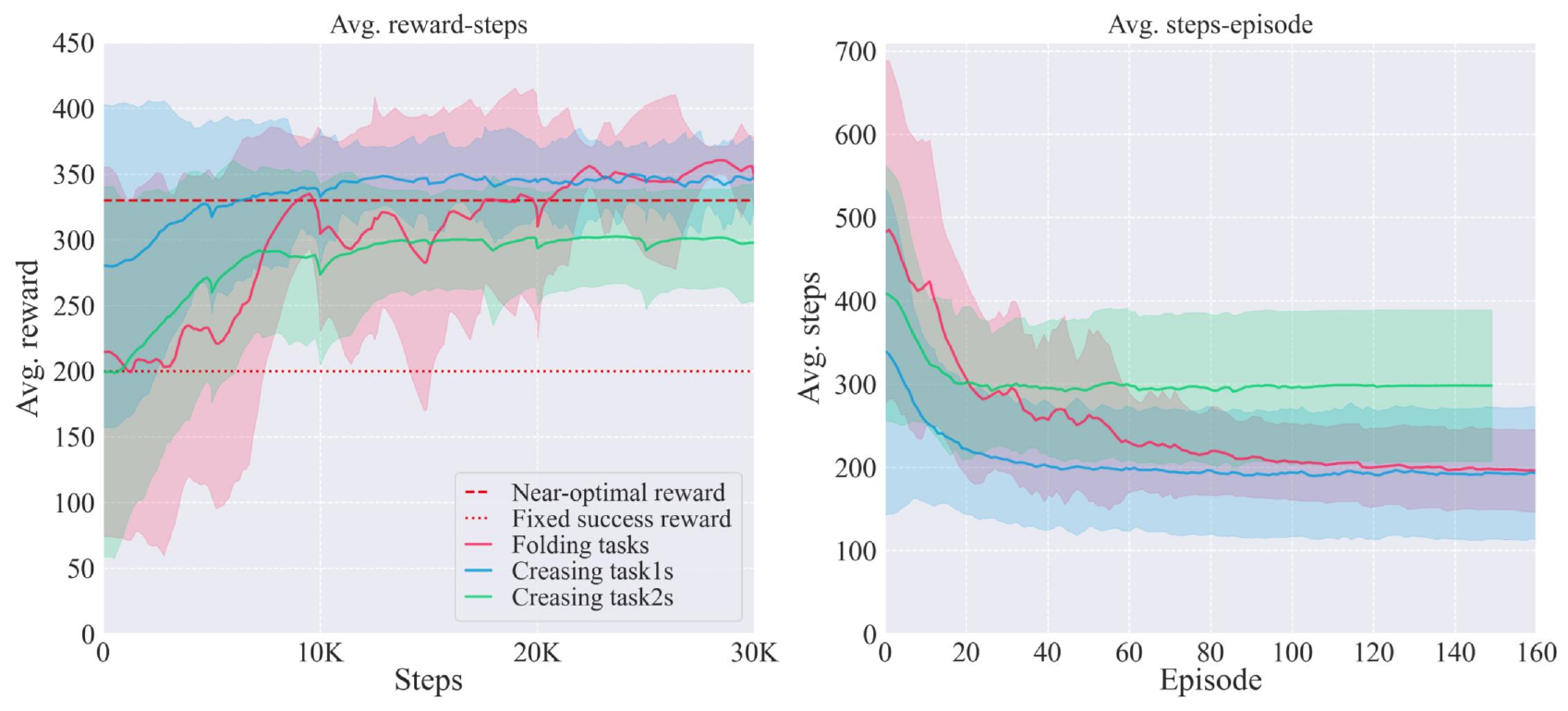}
    \subcaption{\small{Reward and steps in simulation environment}}
    \label{fig:reward_simulation}
    \end{minipage}
    \begin{minipage}[b]{0.5\hsize}
    \centering
    \includegraphics[width=\linewidth]{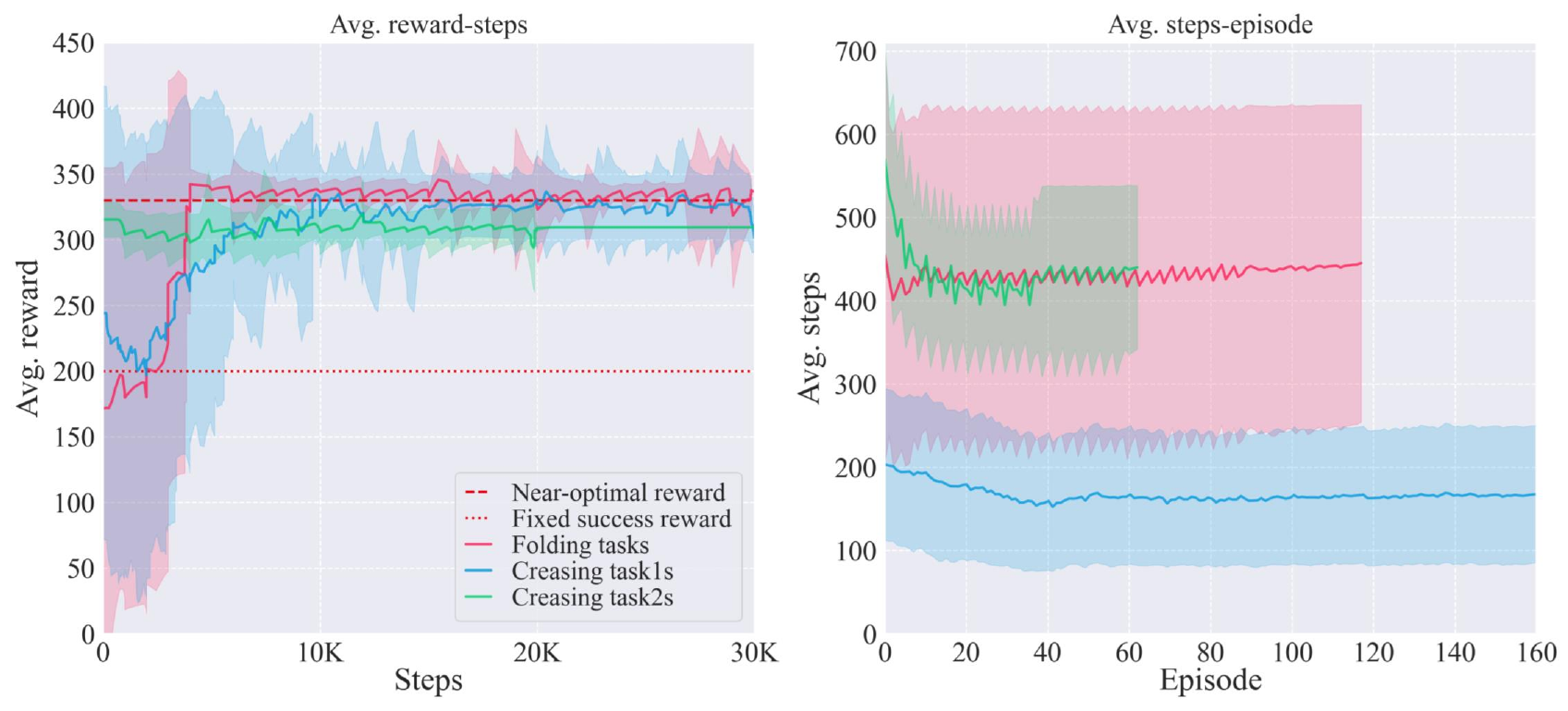}
    \subcaption{\small{Reward and steps in real environment}}
    \label{fig:reward_real}
    \end{minipage}
    \caption{\small{Experimental results}}
        \vspace{-4mm}
    \label{fig:output}
\end{figure*}

The result of wrapping the paper with $0.09\,\mathrm{mm}$ thickness is shown in Fig.~\ref{fig:output}(a). By using the traind policies, the agent was able to realize the entire wrapping process without any tears or wrinkles. The learning curves 
for each task in the simulation and real environments are shown in Figs.~\ref{fig:output}(b) and (c), respectively.

The learning curves are smoothed using an exponential moving average (EMA). The red dashed and dotted lines indicate the optimal reward and the fixed success reward given by the reward function for achieving each task, respectively. Since the cumulative reward values converge near the optimal one, the agent is considered to achieve near-optimal policies in both environments. In addition, the number of steps required to complete each task decreases as the learning progresses, indicating that the agent can accomplish each task more quickly.

\subsection{Performance of Trained Skill Policy}

We compared three metrics—task success rate, tear rate, and wrinkle rate—between the initial and trained policies. Each subtask was executed $10$ times each. As shown in Table \ref{comparison_policy}, the task success rate improved by $82.5\,\mathrm{\%}$ for FT, $77.5\,\mathrm{\%}$ for CT1, and $50.0\,\mathrm{\%}$ for CT2.
If the initial policies are used, the average tears rate in CT1 is $29\,\mathrm{\%}$ and 
average wrinkles rate in FT is $24\,\mathrm{\%}$. 
On the other hand, if the trained policies are used, the tears rate is reduced to almost $1\,\mathrm{\%}$ and the wrinkles rate is reduced to $2.5\,\mathrm{\%}$.

\begin{table}[t]
\captionsetup{textfont={sc,footnotesize}, labelfont=footnotesize, justification=centering, labelsep=newline}
\centering
\caption{\small{Comparison of success rates and tears/wrinkles rates for each task between initial and learned policies}}
\label{comparison_policy}
\begin{tabular}{ccccc}
\hline \hline
\multirow{2}{*}{\textbf{Policy}} & \multirow{2}{*}{\textbf{Task}} & \textbf{Success} & \multicolumn{2}{c}{\textbf{Failure factor}}\\ \cline{4-5}
 & & \textbf{rate} & \textbf{Tears} & \textbf{Wrinkles}\\ \hline
\multirow{3}{*}{\textbf{Initial}} & FT & $5/40$ & $6/40$ & $21/40$ \\ \cline{2-5}
 & CT1 & $15/80$ & $41/80$ & $9/80$ \\ \cline{2-5}
 & CT2 & $19/40$ & $0/40$ & $8/40$ \\ \hline
\multirow{3}{*}{\textbf{Trained}} & FT & $38/40$ & $0/40$ & $2/40$ \\ \cline{2-5}
 & CT1 & $77/80$ & $2/80$ & $1/80$ \\ \cline{2-5}
 & CT2 & $39/40$ & $0/40$ & $1/40$ \\ \hline \hline
\end{tabular}
\end{table}

\subsection{Evaluating Trained Force Controller}

To evaluate the force control policy, we compare the target and actual force transitions in FT-1 and CT1-1. In this experiment, we evaluated the force in $x,y$ and $z$ directions for the paper with medium thickness.
In FT-1, the force is required in the direction of paper tension (positive $y$ and negative $z$ directions) without applying share force in $x$ direction. On the other hand, CT1-1 requires the force to press the paper in negative $y$ direction.
In the green-shaded area, a penalty was applied due to paper tearing during the learning process, while in the yellow-shaded area, no additional penalties were imposed. Although the force generally tracked the target values, there were instances where it deviated, such as between steps 300 and 360 in FT-1. This deviation occurred because the agent was frequently penalized for tearing the paper, prioritizing tear avoidance over strict adherence to the target force.

\begin{figure}
    \centering
    \includegraphics[width=\linewidth]{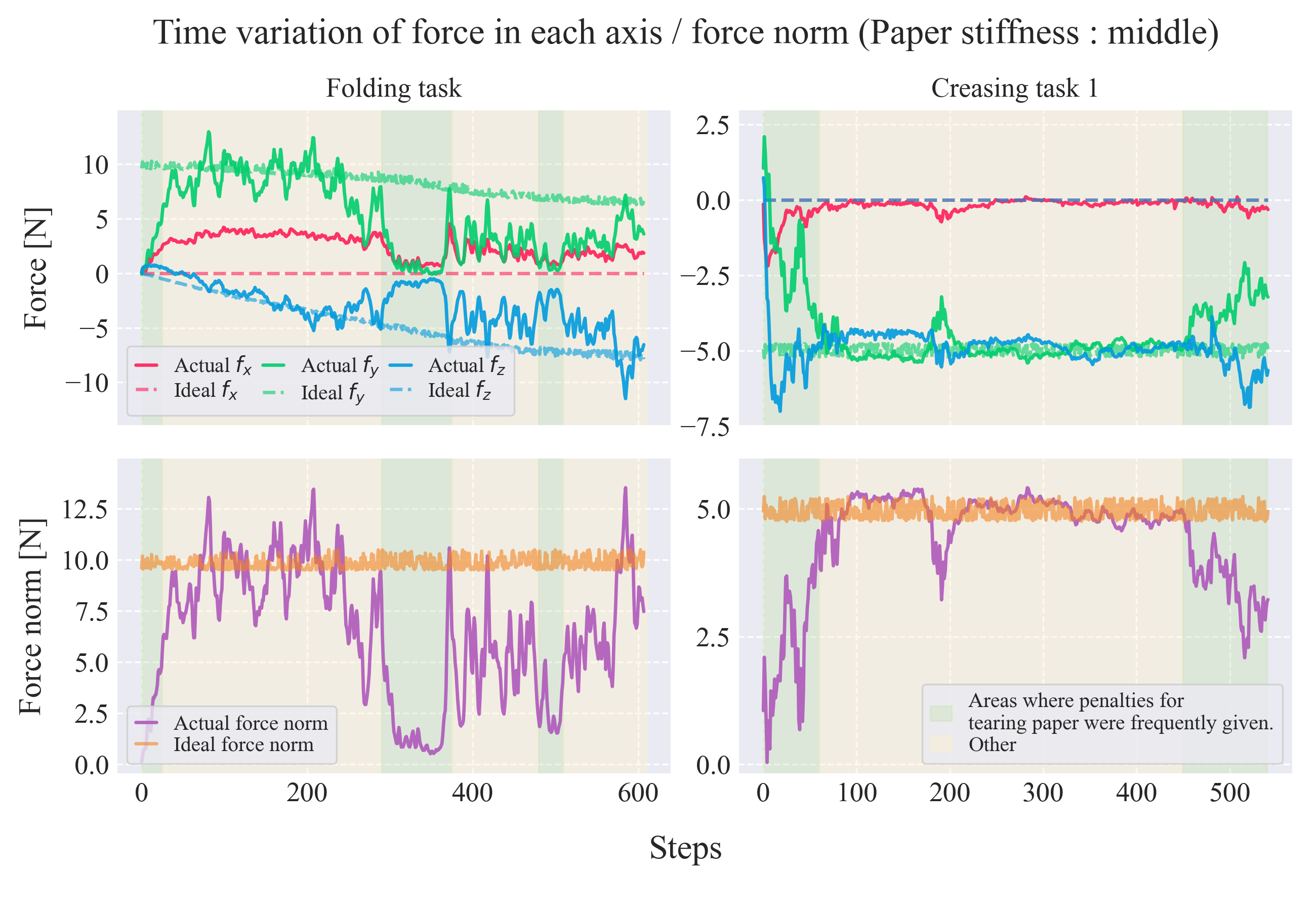}
    \caption{\small{Measured force during each task execution}}
        \vspace{-7mm}
    \label{fig:force_evaluation}
\end{figure}

\subsection{Evaluating Generalizability and Robustness}

We evaluated the generalizability and robustness across varying paper materials and box sizes.

First, we assessed generalizability with respect to changes in paper material, using three types of paper with different thicknesses: high-thickness paper(0.26 $\mathrm{mm}$), which is hard to tear but prone to wrinkling; medium-thickness paper (0.09 $\mathrm{mm}$), offering a balanced resistance to tearing and wrinkling compared to the other two; and low-thickness paper (0.07 $\mathrm{mm}$), which is easy to tear but resistant to wrinkling. The medium-thickness paper is the same as that used in Section \ref{experiments_results}-B. 
With these papers, we compared four metrics: task success rate, $f_r$ acquired by RL, and tears/wrinkles rate. 

For the high- and low-thickness papers, we fine-tuned the policies learned with the medium-thickness paper over approximately 20k steps. In the comparison experiments, each subtask—-comprising FT and CT1—-was performed 10 times for each paper. The results are presented in Table \ref{stiffness_comparison}.
The high success rate indicates that the task can be performed with high accuracy for all paper materials.
In addition, the norm of the desired force $f_r$ indicates that the magnitude of force can be adjusted according to the paper material, and the desired force is learned to be larger for thicker paper and smaller for thinner paper.
\begin{table}[t]
\captionsetup{textfont={sc,footnotesize}, labelfont=footnotesize, justification=centering, labelsep=newline}
\centering
\caption{\small{Task success rates for different paper materials}}
\label{stiffness_comparison}
\begin{tabular}{cccccc}
\hline \hline
\textbf{Paper} & \multirow{2}{*}{\textbf{Task}} & \textbf{Success} & \multirow{2}{*}{\textbf{$\|\bm{f}_{\text{idl}}\|\,\mathrm{[N]}$}} & \multicolumn{2}{c}{\textbf{Failure factor}}\\ \cline{5-6}
\textbf{thickness} & & \textbf{rate} & & \textbf{Tears} & \textbf{Wrinkles} \\ \hline
\multirow{3}{*}{\begin{tabular}{c}\textbf{High} \\ \textbf{$(0.26\,\mathrm{mm})$} \end{tabular}} & FT & 36/40 & 12.41 & 0/40 & 3/40 \\ \cline{2-6} 
& CT1 & 78/80 & 10.15 & 1/80 & 1/80 \\ \cline{2-6} 
& CT2 & 40/40 & N/A & 0/40 & 0/40 \\ \hline
\multirow{3}{*}{\begin{tabular}{c}\textbf{Medium} \\ \textbf{$(0.09\,\mathrm{mm})$} \end{tabular}} & FT & 38/40 & 9.84 & 0/40 & 2/40 \\ \cline{2-6} 
& CT1 & 77/80 & 4.96 & 2/80 & 1/80 \\ \cline{2-6} 
& CT2 & 39/40 & N/A & 0/40 & 1/40 \\ \hline    
\multirow{3}{*}{\begin{tabular}{c}\textbf{Low} \\ \textbf{$(0.07\,\mathrm{mm})$} \end{tabular}} & FT & 31/40 & 1.53 & 4/40 & 4/40 \\ \cline{2-6} 
& CT1 & 61/80 & 3.63 & 12/80 & 7/80 \\ \cline{2-6} 
& CT2 & 34/40 & N/A & 1/40 & 5/40 \\ \hline \hline
\end{tabular}
\end{table}

Second, we evaluated the generalizability of the method for varying box sizes. In this experiment, classification accuracy was assessed in a simulation environment where the box dimensions were varied within the range of ($0.23 \pm 0.08) \times (0.27 \pm 0.08) \times (0.05 \pm 0.04)\,\mathrm{m}$. 
We evaluated the estimation accuracy when each dimension was varied individually within this range, as well as when all dimensions were randomly varied. The accuracy of the phase estimation network is presented in Table \ref{accuracy_PEN_table}. The results indicate that classification accuracy exceeded $90\,\%$ when each parameter was varied individually, demonstrating particularly high robustness to changes in height. However, when all dimensions were varied randomly, classification accuracy was approximately $90\,\%$.

\begin{table}[t]
\captionsetup{textfont={sc,footnotesize}, labelfont=footnotesize, justification=centering, labelsep=newline}
\centering
\caption{\small{Phase estimation accuracy for different papers}}
\label{accuracy_PEN_table}
\begin{tabular}{clcccc}
\hline \hline
\multicolumn{3}{c}{\multirow{2}{*}{\textbf{Parameter [Variation]}}} & \multicolumn{3}{c}{\textbf{Accuracy}} \\ \cline{4-6} 
 & & & FT & CT1 & CT2 \\ \hline
 \textbf{Default} & $\bm{[\pm 0 \mathrm{cm}]}$ & \begin{minipage}{0.8truecm}
      \centering
       \scalebox{0.06}{\includegraphics{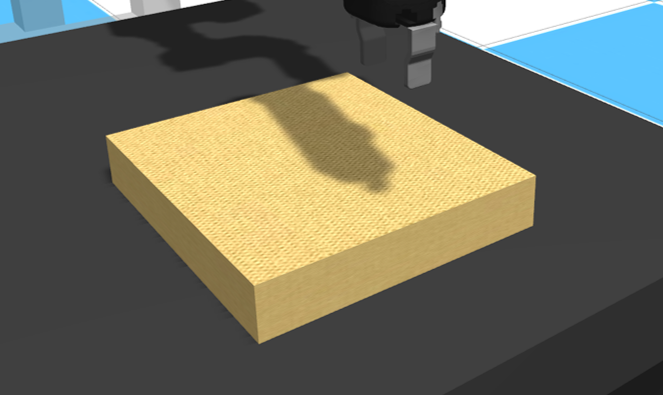}}
    \end{minipage} & $100.0 \%$ & $100.0 \% $ & $100.0 \%$ \\ \hline
\textbf{Length} & $\bm{[\pm 8 \mathrm{cm}]}$ & \begin{minipage}{0.8truecm}
      \centering
      \scalebox{0.06}{\includegraphics{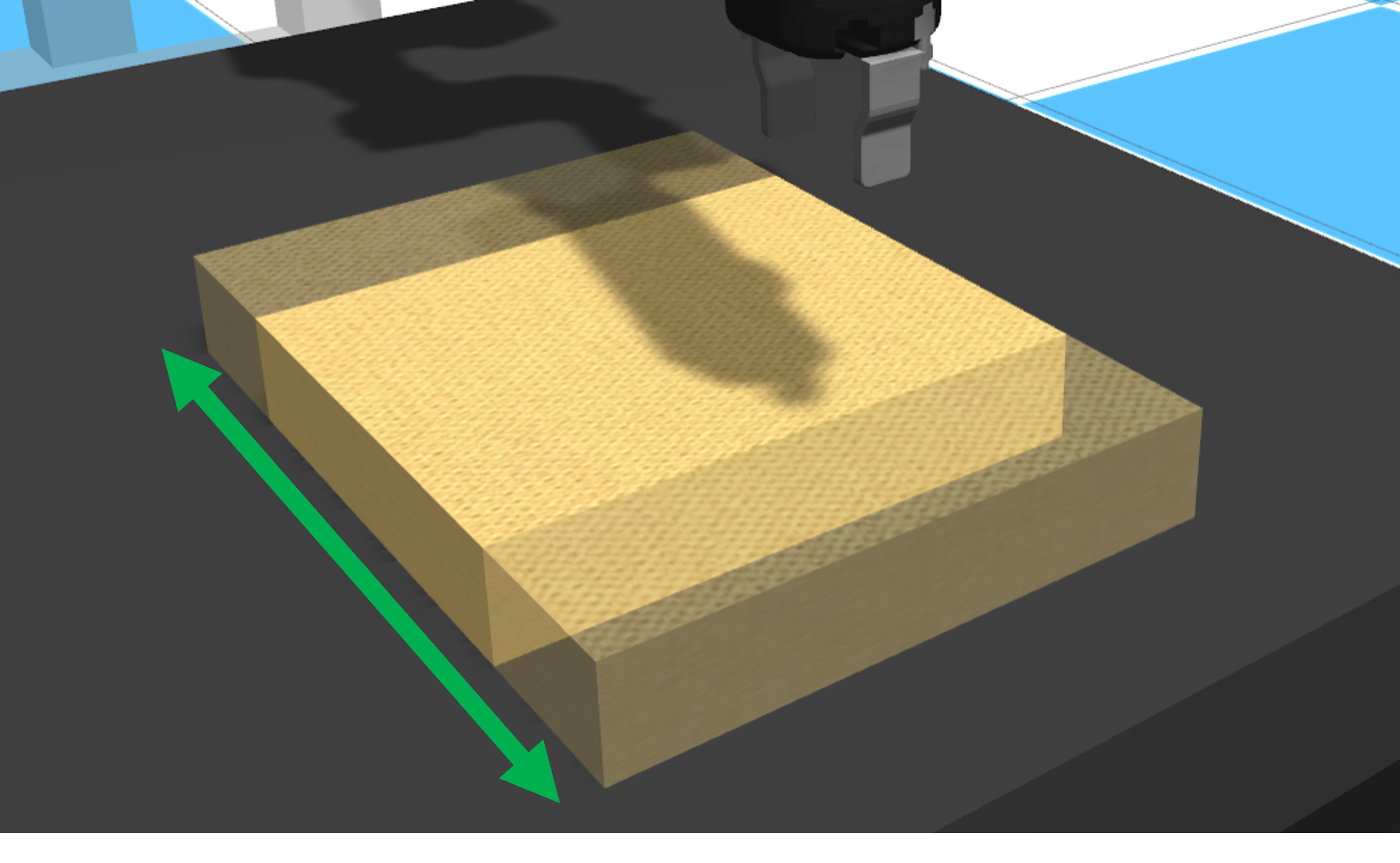}}
    \end{minipage} & $98.9 \%$ & $99.1 \% $ & $96.9 \%$ \\ \hline
\textbf{Width} & $\bm{[\pm 8 \mathrm{cm}]}$ & \begin{minipage}{0.8truecm}
      \centering
      \scalebox{0.06}{\includegraphics{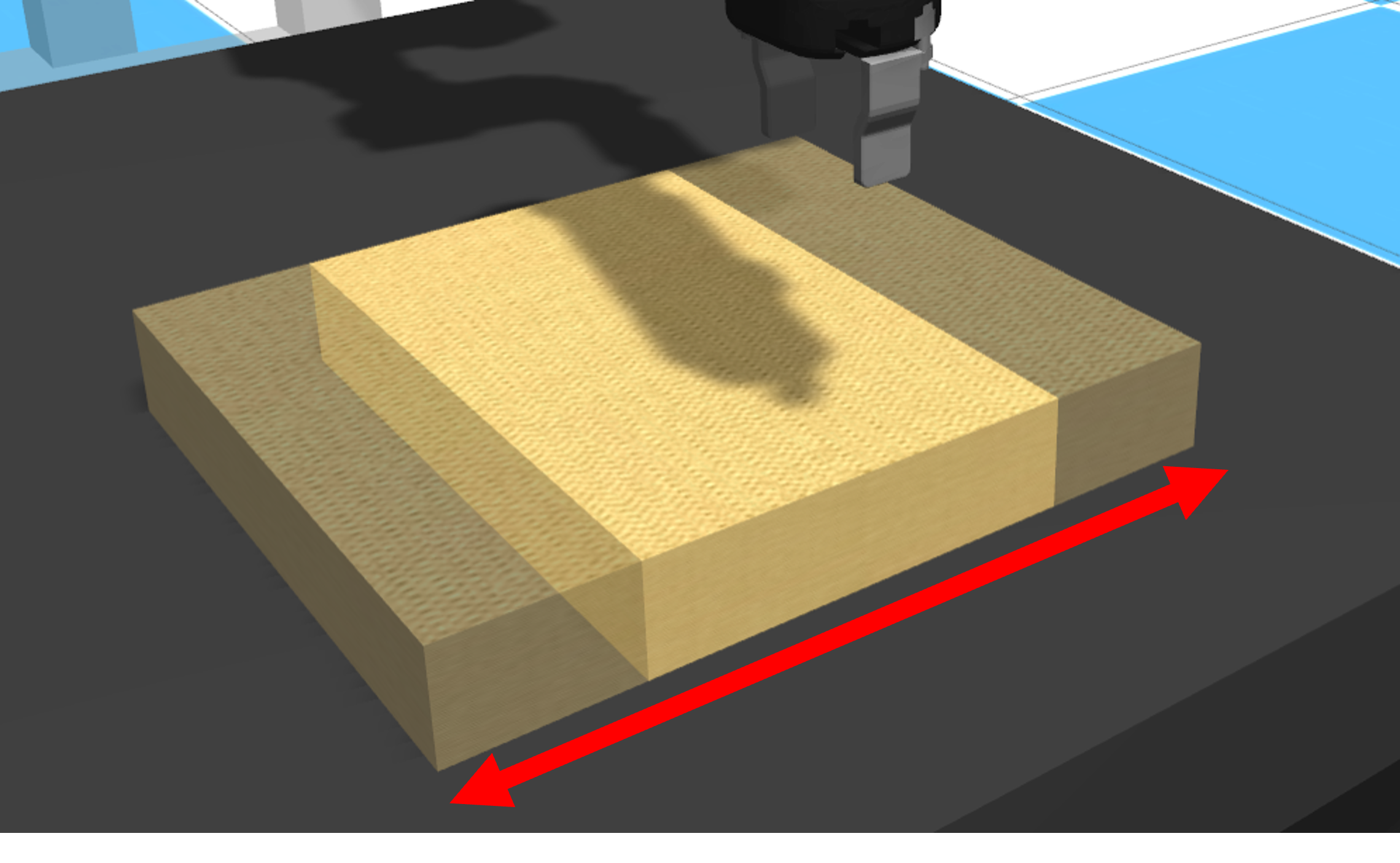}}
    \end{minipage} & $98.6 \%$ & $99.0 \%$  & $97.8 \%$  \\ \hline
\textbf{Height} & $\bm{[\pm 4 \mathrm{cm}]}$ & \begin{minipage}{0.8truecm}
      \centering
      \scalebox{0.06}{\includegraphics{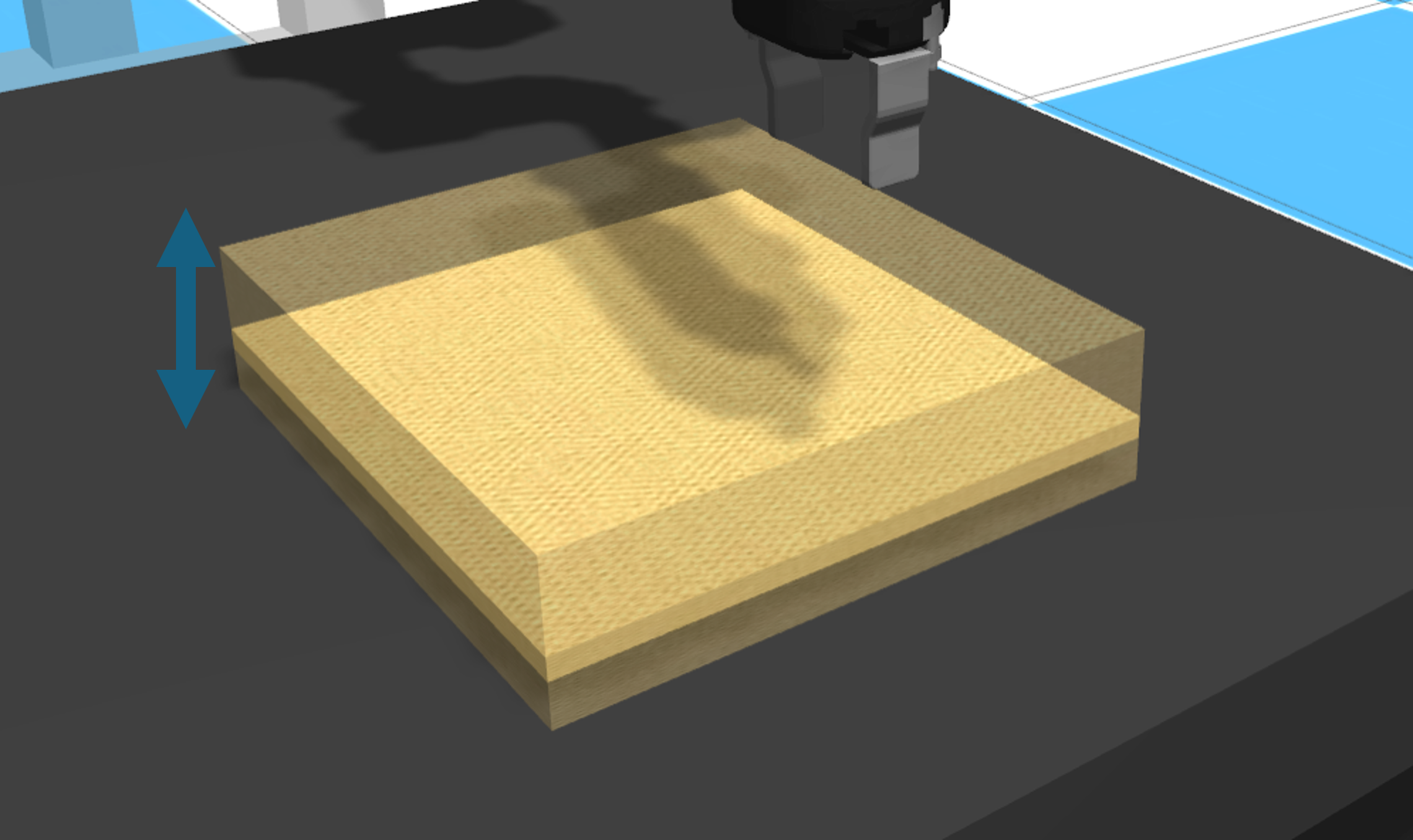}}
    \end{minipage} & $99.8 \%$ & $99.8 \% $ & $99.8 \% $ \\ \hline
\multicolumn{2}{c}{\textbf{All (Random)}} & \begin{minipage}{0.8truecm}
      \centering
      \scalebox{0.06}{\includegraphics{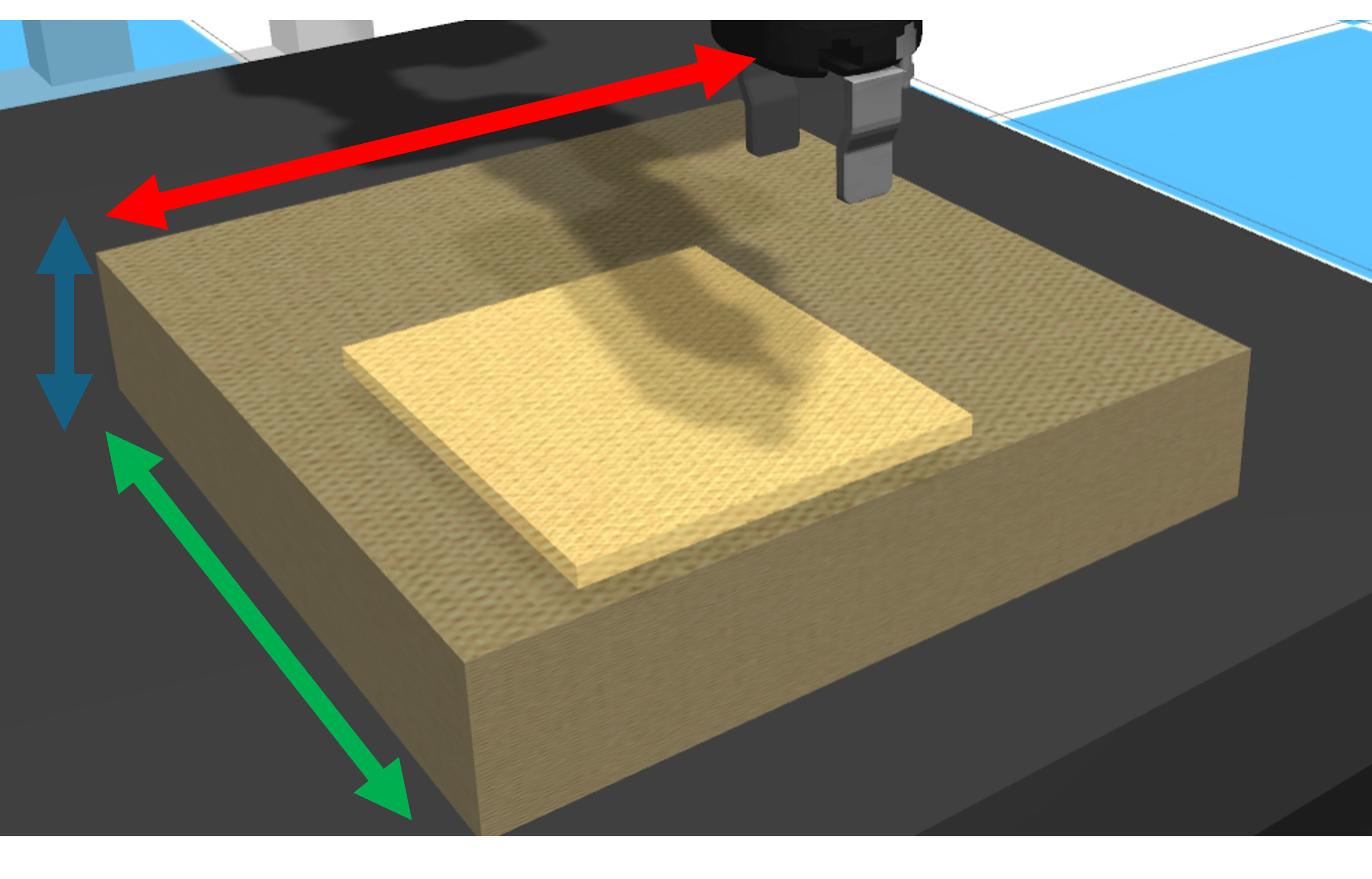}}
    \end{minipage} & $93.7 \%$ & $95.7 \%$ & $87.4 \%$ \\ \hline \hline
\end{tabular}
\end{table}

\subsection{Ablation Study on Learning Strategy}

We compared the proposed method with two alternative approaches: one using only RL (without IL) and another where only position control parameters were learned via RL (without tuning the force control policy). Using medium-thickness paper, we evaluated the task success rate along with the rates of tears and wrinkles. During training, after position control parameters were learned in the simulated environment, the force control parameters were fine-tuned over 20k steps in the real-world environment. For the comparison experiments, FT-1 and CT1-1 were each executed 20 times. The results are presented in Table \ref{proposed_method_comparison}.

When only RL was used, the paper tore or wrinkled more frequently compared to the proposed method, and the task success rate dropped to one-fourth. The paper tore in $17.5\,\%$ of cases and wrinkled in $35\,\%$. Specifically, in FT-1, the paper often flexed and wrinkled due to the linear movement of the TCP from the initial to the target position. In CT1-1, wrinkles occurred frequently due to significant deviations of the TCP's position from the intended target.

When only the position control parameters were trained using RL, the paper tore or wrinkled more frequently than with the proposed method, and the task success rate dropped to one-sixth. Additionally, tearing and wrinkling occurred more frequently. Specifically, the excessive force exerted by the TCP on the paper frequently led to tearing. Conversely, in many instances, insufficient force applied by the gripper resulted in an inability to create creases.

\begin{table}[t]
\captionsetup{textfont={sc,footnotesize}, labelfont=footnotesize, justification=centering, labelsep=newline}
\centering
\caption{\small{Comparison of wrapping task success rate with different learning strategies}}

\label{proposed_method_comparison}
\begin{tabular}{ccccc}
\hline \hline
\multirow{2}{*}{\textbf{Method}} & \multirow{2}{*}{\textbf{Task}} & \textbf{Success} & \multicolumn{2}{c}{\textbf{Failure factor}}\\ \cline{4-5}
& & \textbf{rate} & \textbf{Tears} & \textbf{Wrinkles} \\ \hline
\textbf{Ours} & FT-1 & 19/20 & 0/20 & 1/20 \\ \cline{2-5} 
\textbf{(IL+RL)} & CT1-1 & 18/20 & 2/20 & 0/20 \\ \hline
\multirow{2}{*}{Only RL} 
& FT-1 & 5/20 & 4/20 & 10/20 \\ \cline{2-5} 
& CT1-1 & 4/20 & 5/20 & 5/20 \\ \hline
\multirow{2}{*}{Only PC}    
& FT-1 & 3/20 & 11/20 & 3/20 \\ \cline{2-5} 
& CT1-1 & 3/20 & 9/20 & 6/20 \\ \hline \hline
\end{tabular}
        \vspace{-7mm}
\end{table}
\section{Conclusions}

In this study, we presented a novel approach to robotic wrapping, effectively handling multiple task phases while avoiding wrinkles and tears. By integrating IL and RL, we enabled the robot to learn both nominal trajectories and force control parameters. Experimental results demonstrated the feasibility of executing the wrapping task across various paper materials, with the phase estimation network 
proving robust to variations in box size.
Moreover, ablation studies confirmed that the combination of IL for trajectory learning and RL for force control significantly enhances the success rate of the wrapping process. Moving forward, we aim to expand this system by incorporating visual information, with the goal of achieving more sophisticated wrapping operations in complex environments using dual-arm coordination.

\normalem

\end{document}